# Jointly Identifying Opinion Mining Elements and Fuzzy Measurement of Opinion Intensity to Analyze Product Features

Haiqing Zhang*, Aicha Sekhari*, Yacine Ouzrout*, Abdelaziz Bouras†*

*DISP laboratory, University Lumière Lyon 2, France, 160 Bd de l'Université 69676 Bron Cedex
†Computer Science Department - Qatar University, ictQATAR, Box. 2731, Doha, Qatar
{aicha.sekhari, yacine.ouzrout}@univ-lyon2.fr
abdelaziz.bouras@qu.edu.qa
haiqing.zhang.zhq@gmail.com

**Abstract:** Opinion mining mainly involves three elements: feature and feature-of relations, opinion expressions and the related opinion attributes (e.g. Polarity), and feature-opinion relations. Although many works have emerged to achieve its aim of gaining information, the previous researches typically handled each of the three elements in isolation, which cannot give sufficient information extraction results; hence, the complexity and the running time of information extraction is increased. In this paper, we propose an opinion mining extraction algorithm to jointly discover the main opinion mining elements. Specifically, the algorithm automatically builds kernels to combine closely related words into new terms from word level to phrase level based on dependency relations; and we ensure the accuracy of opinion expressions and polarity based on: fuzzy measurements, opinion degree intensifiers, and opinion patterns. The 3,458 analyzed reviews show that the proposed algorithm can effectively identify the main elements simultaneously and outperform the baseline methods. The proposed algorithm is used to analyze the features among heterogeneous products in the same category. The feature-by-feature comparison can help to select the weaker features and recommend the correct specifications from the beginning life of a product. From this comparison, some interesting observations are revealed. For example, the negative polarity of video dimension is higher than the product usability dimension for a product. Yet, enhancing the dimension of product usability can more effectively improve the product.

**Keywords: opinion mining; dependency relations; fuzzy sets and logic; opinion degree intensifiers; feature-by-feature analysis**

## 1. Introduction

The widely used Web communication on mobile and web-based technologies has dramatically changed the way individuals and communities express their opinions. More and more reviews are posted online to describe customers' opinions on various types of products. These reviews are fundamental pieces of information needed to support both firms and customers to make good decisions. The features and attributes of a product extracted from online customer reviews can be used in recognizing the strengths and weaknesses of the heterogeneous products for firms. While customers do not always have the ability to wisely choose among a variety of products in the market, they commonly seek product information from online reviews before purchasing a new product.

However, the number of reviews grows rapidly, so it becomes impractical to analyze them by hand. In addition, the inherent characteristics of the reviews are diverse and complex. Firms tend to portray the products in different ways, which makes the products more distinguishable and prevents the products from being substituted for each other easily. The heterogeneous products in the same category have slightly different functions, features, and physical characteristics. By doing this, more products become competitive; price alone is not the most important factor to be successful in competing products any more. This internal rule of releasing products causes the number of reviews to increase fast. If the related opinions towards features can be obtained from the massive reviews, the firms will greatly benefit by using the extracted information to evaluate how and where to improve the product through the product development process. Hence, extracting information from the online reviews is academically challenging and has practical use.

Identifying the opinions in a large-scale document of customer reviews is an opinion mining issue, which is a subdivision of information extraction that is concerned with: the features, the opinion it expresses, and the relations between features and opinion expressions. An opinion is a positive or negative sentiment or attitude about an entity or an aspect of the entity from an opinion holder. An opinion is defined as a quintuple in (Liu and Zhang, 2012). We extend the quintuple into a sextuple by adding the relations among features and opinions, which is shown as ($e_i$, $f_{ij}$, $oo_{ijkl}$, $r_{ijkl}$, $h_k$, $t_l$), where $e_i$ is the name of an entity; $f_{ij}$ is a feature of $e_i$; $oo_{ijkl}$ is the opinion expression on feature $f_{ij}$ of entity $e_i$; $r_{ijkl}$ is the sets of feature-opinion relation extraction, feature-feature relation extraction, and opinion-opinion relation extraction; $h_k$ is the opinion holder; and $t_l$ is the time when the opinion is expressed by $h_k$. This definition can provide a basis for transforming unstructured text to structured data. For example, the review presented in Figure 1 has six constitutes: product name ($e_i$), product features ($f_{ij}$), product opinions ($oo_{ijkl}$), relations ($r_{ijkl}$), opinion holder ($h_k$), and opinion post time ($t_{ij}$). The sextuple resolves the unstructured text data into a formalized structured text. The added attribute $r_{ijkl}$ can be used to summarize the overall attitude of the whole review and reflect the opinions with respect to a specific feature. For instance, in the following "Ronald J. Magdos (opinion holder) has only "good" (product opinions) things to say about "Canon Powershot SX510HS's" (entity) in regards to the "photos" (product features) since he has discovered this product<photos good>(relations) on "October 9, 2014" (post time)" will obtain parameters e, f, h, t, oo, and r based on the indices i=1, j=1,k=1, and l=1.

⭐⭐⭐⭐⭐ **I wants something simple to use that takes good photos, this camera does both**, October 9, 2014
By <u>Ronald J. Magdos</u> (DeMotte, IN USA) - <u>See all my reviews</u>
REAL NAME
Verified Purchase (What's this?)
This review is from: Canon PowerShot SX510 HS 12.1 MP CMOS Digital Camera (discontinued by manufacturer) (Electronics)
I wants something simple to use that takes good photos, this camera does both. The zoom is GREAT, and the wifi feature is very user friendly. I would recommend this camera to anyone who wants ease to use, good photos, and a great price.

Opinion extraction presented in a sextuple

| $e_i$ | (Canon PowerShot SX510 HS 12.1 MP CMOS Digital Camera)$_i$ | | | |
|---|---|---|---|---|
| $f_{ij}$ | (photos; camera; zoom; wifi; feature; camera; photos; price)$_j$ | | | |
| $oo_{ijkl}$ | (simple; good; great; very; friendly; recommend; easy; good; great) | | | |
| $r_{ijkl}$ | feature-feature | Opinion-opinion | feature-opinion | |
| | <wifi feature> | <simple to use> <very user friendly> <easy to use> | <camera          simple to use> <photos          good> <zoom          great> <wifi feature          very user friendly> <camera          recommend          easy to use> <photos          good> <price          great> 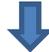 | <camera$^2$          simple to use          recommend          easy to use > <photos          good>$^2$ <zoom          great> <wifi feature          very user friendly> <price          great> |
| $h_k$ | (Ronald J. Magdos)$_k$ | | | |
| $t_l$ | (October 9, 2014)$_l$ | | | |

Figure 1. Extracted opinions and features showing in a sextuple

In sextuple, the features $f_{ij}$, opinion expression $oo_{ijkl}$, and opinion description $r_{ijkl}$ are the necessary pieces of information which are difficult to obtain. Two fundamental problems of mining such information are opinion features extraction and opinion words locating.

1)  Opinion features are characteristics of the products in which the opinion has been described. Two issues are generated in product feature extraction.
    a)  ***One is that synonyms are often occurring in extraction of features.*** For example, 'image' and 'photo' refers to the same product feature in camera reviews.
    b)  ***The other one is some product features are combined by several nouns***. The obtained 'wifi feature' in Figure 1 is an example. Hence, feature-of relation is used to record the synonyms of features and rebuild the noun terms to more accurately represent product features.
2)  Opinion expressions are the opinion words that the reviewers have adopted to describe their opinions on the related features.

a) ***Opinion expressions are commonly composed by an opinion pattern involving adjectives, adverbs, and verbs instead of a single opinion word.*** Thus, opinion-of relation extraction is adopted to keep the opinion patterns. For example, 'simple to use', 'very user friendly', and 'easy to use' in Figure 1.

b) ***Opinion expressions also need to express the evaluation for correct targets.*** For example, "I had a nice trip to Yellowstone. I took some very nice pictures and videos by using this camera". "nice" describes the 'trip', but "trip" is not the feature of a digital camera. Therefore, "nice" does not need to be extracted in this case.

c) The feature-opinion relation extraction is necessary to be proposed to express the opinion expressions corresponding with the related opinion features.

An interesting observed phenomenon from the obtained sextuple is that the degree of success about the heterogeneous products in the same category is vastly different. Some features are key impact factors and have a great influence on decisions for purchasing (customers), as well as product development strategies (firms). Reducing the weakness of important features can increase consumption quantities and enhance the product's reputation effectively. A straightforward solution for important, but weaker, feature identification is to select all of the important features that have the most negative comments as the weak features. However, customers' opinions on the worst features may not have the greatest influence on their overall opinions, and thus will not influence the general consumption quantities. Moreover, a feature's weakness may affect other features, causing negative comments, emphasis, and improvements to be focused on the features that are not the root cause of the issue. Thus the worst features could lead to firms spending more effort into maintaining features and cannot gain the maximum benefits from them. Hence, a comprehensive analysis strategy for feature-by-feature comparison has to be able to identify the important features that are currently weak.

This work identifies the strengths and weaknesses of heterogeneous products in the same category; selects the most successful type as the benchmark product; then recommends improvements for the appropriate weaker features of the weaker products. The aim is to make more heterogeneous products more successful during the on-shelf time, so that the same brand will be more successful. To be able to achieve this aim, we decompose the problem of information extraction into the following tasks.

1). Extraction fundamental information: 1.1). Mining products' features that have been created by the reviews; identifying opinion expressions on each review; building feature-opinion relations. 1.2). Concentrate on representing the correct results of feature-opinion relation as it contains products' features and opinion expressions. 1.3). Measure the negative and positive polarity of opinion phrases obtained from feature-opinion relations to calculate reviewers' overall orientation.

2). Cases application: Identify the strengths and the weaknesses of the heterogeneous products in the same category based on the obtained results of sentiment analysis in task 1. Predict the future polarity tendencies of weaker products when setting new values for its weaker features. Finally, propose the features that need to be improved to increase the benefits for the firms.

The remainder of this paper is organized as follows. Section 2 reviews the main related works. Section 3 represents the mechanism of assigning the polarity and intensity of opinion expressions. Section 4 introduces the opinion mining algorithm to jointly complete opinion mining tasks. Section 5 conducts the experiments in multiple aspects and analyzes the superiorities and the deficiencies of the proposed algorithm by comparing with the baseline works. Section 6 concludes the paper with future perspectives.

## 2. Related work

In this paper, we focus on jointly detecting the three principle elements in the reviews: feature and feature-of relation, opinion and opinion pattern extraction, and feature-opinion extraction. In previous works, these elements have mostly been studied in isolation. Therefore, we treat these three elements as three separate tasks and study the related works.

The existing works on feature extraction can be divided into three groups: frequent term mining, supervised sequence labeling, and unsupervised and knowledge-learning based approach. The most representative work for "frequent term mining" approach is Hu and Liu (2004), which is restricted to detecting the features that are strongly associated with a single noun and considers only adjectives collocated with the near feature words as opinion expressions. Some additional works (Popescu and Etzioni, 2007; Zhuang et al., 2006; Qiu et al., 2011; Blair-Goldensohn et al., 2008) involve manually constructed rules and semantic analysis, but these still cannot fully reduce the disadvantages of this branch. The "supervised sequence labeling" (Jakob and Gurevych, 2010; Choi and Cardie, 2010; Wu et al., 2009; Jin and Ho, 2009) usually needs a large amount of training data that is mainly com-posed by hand-labeled training sentences. All of the methods mentioned above do not have the ability to group semantically related expressions together. The existing works belonging to "unsupervised and knowledge-learning based approach" (topic modeling) are based on two models: PLSA (Probabilistic Latent Semantic Analysis) (Hofmann, 1999) and LDA (Latent Dirichlet Allocation) (Blei et al., 2003). According to the work(Titov and McDonald, 2008), the existing models are not suitable to be used to detect features, because they can only work well for capturing global topics, but cannot intelligently understand human judgments. Some excellent models (Brody and Elhadad, 2010; Mukherjee and Liu, 2012) have emerged to overcome the drawbacks.

The opinion expressions consist of a set of opinion words, which are used to present the polarities of sentiments and measure the strength of the expressed opinions. Previous research can be divided into two main categories: CRF (Conditional Random Field)-based approaches and parsing-based approaches. Most of the CRF-based approaches mainly focus on one direction and single word expressions. For instance, Breck et al. (2007) formulated the opinion expression extraction as a token-level sequence labeling problem and proposed a CRF-based approach; Choi and Cardie (2010) jointly determine the opinion expressions, polarities, and the strength based on the hierarchical parameter sharing technique. However, all of the approaches belonging to this category are token-level and cannot efficiently extract phrase-level information. Although semi-CRFs (Sarawagi and Cohen, 2004;Okanohara et al., 2006) are proposed to allow sequence labeling in phrase-level, these methods are known to be difficult to implement (Yang and Cardie, 2012). Previous works (Bethard et al., 2006; Kobayashi et al., 2007; Joshi and Penstein-Rosé, 2009; Wu et al., 2009) show that adopting syntactic parsing features to identify opinion expressions and the related attributes is more helpful than the CRF-based approach. Recently, some excellent methods are proposed to conquer the existing limitations. For example, Rill et al. (2012) proposed an approach to generate the lists of opinion bearing phrases based on phrase extraction strategies. This work only adopts the review titles and the star ratings to calculate the opinion values. In general, the obtained opinion values of this work can correctly reflect the degree of users' opinions. However, this work cannot assign reasonable values for some words with only weak polarities and some words with strong negative polarities, because the idea of projecting the opinion values into a one to five scale may not always be true. Johansson and Moschitti (2011) studied the implementation of end-to-end systems for the tasks of opinion expressions extraction by employing dependency relations. This method has shown that by using the joint model through use of features describing the interaction of opinions with linguistic structures significantly outperforms the sequential approach. Yang and Cardie (2012) extend the semi-CRF model proposed by Sarawagi and Cohen (2004) for extracting opinion expressions by considering the syntactic parsing structure during the learning and inference process, which allows opinion expressions to be organized at the phrase-level. Experiments show that the performance of this method is better than the current works.

Moreover, some combination approaches(Zhao et al., 2010; Brody and Elhadad, 2010; Kobayashi et al., 2007) are proposed by considering the impacts between some internal elements. Zhao et al. (2010) jointly capture both feature and opinion expressions within topic models by extending existing topic models. Brody and Elhadad (2010) proposed an unsupervised system for feature extraction and sentiment determination in each review with consideration to the influence of features on opinion polarity. Kobayashi et al. (2007) defined the opinion unit as a quadruple. In this work, feature-opinion relations and feature-of relations are obtained based on the methods of combining contextual and statistical clues. In conclusion, all of the approaches have their own advantages and disadvantages. Although some models obviously outperform others in each element, to the best of our knowledge, *there is no solution that is simultaneous proficient in all three elements in practice.* In the opinion mining processes, the three elements usually lie in a labyrinth of relationships and one element will encounter another element in each sentence, which makes the opinion mining results directly unobtainable. To be able to gain more benefits from real-life application for firms and customers, we aim to find a *compromising solution that allows the three elements to be taken into account as an integrated unit instead of seeking the best approach for one element.*

## 3. Fuzzy weights assigning for opinion expressions

### 3.1 Opinion Pattern Extraction

This section will propose the methods to calculate the orientations of opinion words. The existing patterns of opinions or opinion combinations are proposed and the fuzzy weights of opinion patterns are given. The work will use the part-of-speech (POS) tool proposed by Stanford[1](Toutanova et al., 2003) to assign parts of speech for each word, which is the process of marking up a word in a text as corresponding to a particular part of speech (noun, verb, ad-jective, etc.) based on its definition and context.

---

[1] http://nlp.stanford.edu/software/tagger.shtml

Not all the words in review sentences are useful for identifying product features and orientations of the discussed product. Nouns and noun phrases in the sentences are considered as features; adjectives are considered as opinion expressions in previous works (Hu and Liu 2004; Qiu et al., 2011; Kar and Mandal, 2011; Liu, 2012). But some verbs and adverbs can also express opinions, such as: 'like', 'recommend', 'appreciate', and so on. Thus we extend the potential POS tags for opinion words to JJ (adjectives), VB (Verbs), and RB (adverbs). Adjectives, adverbs, or verbs, are good indicators for finding opinions. However, an isolated adjective/adverb/verb cannot determine the opinion orientation or intensity with insufficient context. Moreover, a single word cannot completely express the related opinion or feature. Therefore, this work focuses on finding the necessary phrase structures (patterns) that can express the corresponding opinions and features.

The existing works are proposed to perform classification based on fixed syntactic phrases that are likely to be used to express opinions (Turney, 2002; Liu, 2010) by extracting two consecutive words. It should be noted that the third word followed by the two consecutive words is still often used to express the expected features or the intensity of opinion polarities. For example, *This camera produces high quality pictures,* the third word *pictures* is the opinion feature. These three words give a full answer of two questions: why is this good camera and how good the camera is. Take another example: *This is a truly amazing little camera*, the key information will be lost if only truly amazing is extracted, because the words *amazing* and *little* play an equally important role for the feature *camera*. We have deleted the fourth pattern in Turney's work, since pattern (NN/NNS, JJ) is rarely found in the text, and add new patterns that contain verbs based on previous discussions. Table 1 gives the expected phrase extraction patterns. In our work, a max of three consecutive words is recognized as a unit that can express the full opinion of a phrase if their POS tags conform to any of the patterns in Table 1. For example, pattern_1.1 includes three consecutive words if the first word is an adjective, the second word is a noun, and the third word is also a noun.

**Table 1**. Expected opinion phrases pattern

| Pattern_ID | First word | Second word | Third Word | Example |
|---|---|---|---|---|
| 1.1 | JJ | NN/NNS | NN/NNS | good picture quality for an admittedly undiscerning eye.(JJ+NN+NN) <br> The camera has great picture.(JJ+NN) |
| 1.2 | JJ | NN/NNS | RB | My old camera took terrible pictures indoors and in low light.(JJ+NNS+RB) |
| 1.3 | JJ | RB | JJ | This is a great fully functional camera.(JJ+RB+JJ) |
| 2.1 | RB | JJ/RB | NN/NNS/-- | This is extremely good picture.(RB+JJ+NN) <br> The first weekend I took over 300 photographs and I am extremely happy with this camera. (RB+JJ) |
| 2.2 | RB | JJ/RB | JJ | This is a truly amazing little camera. (RB+JJ+JJ) <br> The camera was decidedly too expensive. (RB+RB+JJ) |
| 2.3 | RB | VB +DET | NN/NNS | My sister really like this camera. (RB+VB+DET+NN) |
| 3.1 | VB | RB | JJ | This camera works very great. (VB+RB+JJ) |
| 3.2 | VB | JJ | -- | It works great for a kindle camera. (VBZ+JJ) |

Note 1: we use JJ to express JJ, JJR, and JJS for short; use RB to express RB, RBR, and RBS; use VB to express all types of verb inflection.
Note 2: JJ (adjectives); JJR (comparative adjectives); JJS (superlative adjectives); NN (singular nouns); NNS (plural nouns); RB(Adverb);RBR(comparative adverb);RBS(superlative adverb);
Note 3: we also consider a few words that are tagged by NN as the opinion words. Take 'problem' as an example.

## 3.2 Opinion Intensity Determine

Information lifecycle systems usually rely on the customers or designers to input the data to figure out how the business should progress. The reviews from customers or designers that tend to have the disorganized, vague, and unanalyzed characteristic are often represented by an unstructured pattern. The paper focuses on proposing a common methodology to gather and parse data for inputting more understandable and mature data into information lifecycle systems.

In order to know the orientation of the opinions, the opinion lexicon needs to be selected and collected. Kar and Mandal (2011) provided a seed set of only 20 adjectives and adverbs as well as around 10 verbs to qualify features, which is too weak to completely reflect and apprehend opinions accurately. Around 6,800 positive and negative English opinion words were compiled by Hu and Liu (2004). We have extended these opinion words by adding some words that can express the degree of intensity in the customer's emotion. For example: *'very', 'extremely',*

*'really', 'absolutely',* and etc. We have collected 62 adverbs that are called **Opinion Degree Intensifiers,** which can be used in both positive and negative situations to express the opinion degree or to change the orientation of the opinion. Opinion Degree Intensifiers are grouped into two types: adverbs that only change the opinion degree; and adverbs that will change the orientation of the opinion (Table 3). The opinion expressions have the characteristics of uncertainty as different customers will adopt different words to express the same opinion and the same word has different opinion intensities under different circumstances. Fuzzy logic is a sophisticated approach to tackle uncertain and inaccurate issues (Zhang et al., 2014). Therefore, five fuzzy degrees are defined for the first type of words based on the intensity of the adverbs. Three fuzzy degrees are given for the second type of words, because there are fewer words that have such function and the gaps among these words are narrow.

The 6800 opinion words are updated with assigned weights that lie in [-1, 1]. The sets of opinion words are categorized into five levels based on the orientation of the word (Table 2). Some words are defined as the benchmark (core), which can be used as the standard when determining the other words' polarities. In Table 2 and Table 3, "L" is the lower value, "M" is the middle value, and "U" is the upper value for the fuzzy element of each defined degree.

**Table 2**. Fuzzy measure of opinion words (positive+negative)

| positive | degree (+) | | | degree (++) | | | degree (+++) | | | degree (++++) | | | degree (+++++) | | |
|---|---|---|---|---|---|---|---|---|---|---|---|---|---|---|---|
| Fuzzy Scale | L | M | U | L | M | U | L | M | U | L | M | U | L | M | U |
| | 0 | 0.1 | 0.3 | 0.1 | 0.3 | 0.5 | 0.3 | 0.5 | 0.7 | 0.5 | 0.7 | 0.9 | 0.7 | 0.9 | 1 |
| Negative | degree (+) | | | degree (++) | | | degree (+++) | | | degree (++++) | | | degree (+++++) | | |
| Fuzzy Scale | L | M | U | L | M | U | L | M | U | L | M | U | L | M | U |
| | -0.3 | -0.1 | -0 | -0.5 | -0.3 | -0.1 | -0.7 | -0.5 | -0.3 | -0.9 | -0.7 | -0.5 | -1 | -0.9 | -0.7 |

**Table 3**. Fuzzy measure of Opinion Degree Intensifiers

| Adverbs (only change the degree) | degree (+) | | | degree (++) | | | degree (+++) | | | degree (++++) | | | degree (+++++) | | |
|---|---|---|---|---|---|---|---|---|---|---|---|---|---|---|---|
| Fuzzy Scale | L | M | U | L | M | U | L | M | U | L | M | U | L | M | U |
| | 0 | 0.1 | 0.3 | 0.1 | 0.3 | 0.5 | 0.3 | 0.5 | 0.7 | 0.5 | 0.7 | 0.9 | 0.7 | 0.9 | 1 |
| Adverbs (change orientation of the opinion) | degree (+) | | | degree (++) | | | degree (+++) | | | -- | | | -- | | |
| Fuzzy Scale | L | M | U | L | M | U | L | M | U | -- | -- | -- | -- | -- | -- |
| | -0.5 | -0.3 | -0 | -0.7 | -0.5 | -0.3 | -0.9 | -0.7 | -0.5 | -- | -- | -- | -- | -- | -- |

We obtain the weight of the extracted opinionated phrases as a combination of the weight of individual words in the phrases. The weights of opinion phrases of all patterns are considered as the combination of RB/RBR/RBS, JJ/RB/RBR/RBS, and VBN/VBD. For example, the opinion of pattern 1.1 and 1.2 is determined by the first word (JJ) that can directly be obtained based on the defined opinion words' weights in Table 2. To be able to know the fuzzy weights of every reviewer, two different cases are defined based on different combinations of opinion words in the proposed patterns.

**Definition 3.2.1** (weights of case 1) The opinion is the combination of the opinion degree intensifiers and 6800 opinion words that include adjectives and verbs. The weights of the opinion in case 1 are defined in three types of situations based on the words' orientation, which is shown in the following equation:

*weights of* $\left[ \left( \text{RB/RBR/RBS} \right) combination with \left( \text{JJ/RB/RBR/RBS} \right) \right]$

$$
= \begin{cases}
1. \text{degree} \left( \text{RB/ RBR/ RBS} \right) \oplus \text{degree} (\text{JJ/ RB/ RBR/ RBS}), \\
\quad if \ \text{degree} \left( \text{RB/ RBR/ RBS} \right) > 0 \ \& \ \text{degree(JJ/RB/RBR/RBS)} > 0 \ \text{eg: very good, extremely high} \\
\quad elseif \ \text{degree} \left( \text{RB/ RBR/ RBS} \right) < 0 \ \& \ \text{degree(JJ/RB/RBR/RBS)} > 0 \ \text{eg: not good, not high} \\
2. - \text{degree} \left( \text{RB/ RBR/ RBS} \right) \oplus \text{degree} (\text{JJ/ RB/ RBR/ RBS}), \\
\quad if \ \text{degree} \left( \text{RB/ RBR/ RBS} \right) > 0 \ \& \ \text{degree(JJ/RB/RBR/RBS)} < 0 \ \text{eg: very bad, extremely annoyed} \\
3. \text{degree} \left( \text{RB/ RBR/ RBS} \right) \otimes \text{degree} (\text{JJ/ RB/ RBR/ RBS})), \\
\quad if \ \text{degree} \left( \text{RB/ RBR/ RBS} \right) < 0 \ \& \ \text{degree(JJ/RB/RBR/RBS)} < 0 \ \text{eg : not bad, not annoyed}
\end{cases}
\tag{1}
$$

**Remark 1:** Some example words and their respective weights are calculated by Eq.1 and shown in Table 4. Where degree(very)=(0.5,0.7,0.9), degree(extremely)=(0.7,0.9,1), and degree(not)= (-0.7,-0.5,-0.3) are defined opinion degree intensifiers.

**Remark 2:** All of the positive weights lie in [0, 1] and all of the negative weights lie in [-1, 0]. If any value is beyond this range, then this value equals the boundary value.

**Remark 3:** When the opinion words are the connection between RB/RBR/RBS and VBN/VBD, Eq.1 is still used to calculate the opinion weights.

**Table 4.** Examples of fuzzy measure of opinion phrases for case 1

| Adverbs (only change the degree) | Good | (0.3,0.5,0.7) | Very good | (0.8,1,1) |
|---|---|---|---|---|
| | high | (0.5,0.7,0.9) | Extremely high | (1,1,1) |
| | Bad | (-0.7,-0.5,-0.3) | Very Bad | (-1,-1,-0.8) |
| | annoyed | (-0.5,-0.7,-0.9) | Extremely annoyed | (-1,-1,-1) |
| Adverbs (will change the orientation of the opinion) | Good | (0.3,0.5,0.7) | Not good | (-0.4, 0, 0.4) |
| | high | (0.5,0.7,0.9) | Not high | (-0.2, 0.2, 0.6) |
| | bad | (-0.7,-0.5,-0.3) | Not bad | (0.09, 0.25, 0.49) |
| | annoyed | (-0.5,-0.7,-0.9) | Not annoyed | (0.15,0.35,0.63) |

**Definition 3.2.2** (weights of case 2) Some opinion words appear together with case 1. For instance, "not a very good camera", "extremely high quality", etc. The opinion phrases of such types are calculated by Eq.2 and Eq.3.

$$weights\ of\ \Big[\left(not\,/\,never\,/\,...\right)\Big[combination\ with\left(RB/RBR/RBS\right)combination\ with\left(JJ/RB/RBR/RBS\right)\Big]\Big]$$

$$=\begin{cases}1.\ degree\left(\left(RB/RBR/RBS\right)combination\ with\left(JJ/RB/RBR/RBS\right)\right)\oplus\left(-\left(-degree\left(not\,/\,never\,/\,...\right).^\wedge 2\right)\right)\\ if\ degree(JJ/RB/RBR/RBS)>0\ \ eg:not\ very\ good,\ not\ extremely\ high\\ 2.\ degree\left(\left(RB/RBR/RBS\right)combination\ with\left(JJ/RB/RBR/RBS\right)\right)\oplus\left(\left(-degree\left(not\,/\,never\,/\,...\right).^\wedge 2\right)\right)\\ if\ degree(JJ/RB/RBR/RBS)<0\ \ eg:not\ very\ bad,\ not\ extremely\ annoyed\end{cases} \quad (2)$$

$$weights\ of\ \Big[\left(very\,/\,so\,/\,...\right)\Big[combination\ with\left(RB/RBR/RBS\right)combination\ with\left(JJ/RB/RBR/RBS\right)\Big]\Big]$$

$$=\begin{cases}1.\ degree\left(\left(RB/RBR/RBS\right)combination\ with\left(JJ/RB/RBR/RBS\right)\right)\oplus\left(\left(degree\left(very\,/\,so\,/\,...\right).^\wedge 2\right)\right)\\ if\ degree(JJ/RB/RBR/RBS)>0\ \ eg:very\ very\ good,\ so\ extremely\ high\\ 2.\ degree\left(\left(RB/RBR/RBS\right)combination\ with\left(JJ/RB/RBR/RBS\right)\right)\oplus\left(-\left(degree\left(very\,/\,so\,/\,...\right).^\wedge 2\right)\right)\\ if\ degree(JJ/RB/RBR/RBS)<0\ \ eg:very\ very\ bad,\ so\ extremely\ annoyed\end{cases} \quad (3)$$

Table 5. Examples of fuzzy measure of opinion phrases for case 2

| Good | (0.3,0.5,0.7) | Very good | (0.8,1,1) | not very good | (0.31,0.75,0.91) |
|---|---|---|---|---|---|
| | | | | very very good | (1,1,1) |
| high | (0.5,0.7,0.9) | Extremely high | (1,1,1) | not extremely high | (0.51,0.75,0.91) |
| | | | | so extremely high | (1,1,1) |
| Bad | (-0.7,-0.5,-0.3) | Very Bad | (-1,-1,-0.8) | not very bad | (-0.91, -0.75, -0.31) |
| | | | | very very bad | (-1,-1,-1) |
| annoyed | (-0.9,-0.7,-0.5) | Extremely annoyed | (-1,-1,-1) | not extremely annoyed | (-0.91,-0.75,-0.51) |
| | | | | so extremely annoyed | (-1,-1,-1) |

**Remark 4:** Table 5 gives some examples of case 2 and the corresponding fuzzy weights. The final extracted opinion phrase is the combination of opinion words and the related opinion features. The final fuzzy weights of opinion words are calculated by Eq.(1-3).

**Definition 3.2.3** (Weight for a review). The weight of a review is calculated based on fuzzy operation. How frequently the opinion features appear and the related fuzzy weights of opinion words are two important elements that can determine the weight of a review.

$$RW=\frac{\sum_{i=1}^{n}\text{fuzzy scale}\left(opinion\ words\right)\otimes f\left(\text{Related features}\right)_i}{\sum_{i=1}^{n}f\left(\text{Related features}\right)_i}\ \ (where,\ n\ is\ the\ total\ number\ of\ features\ in\ a\ review)\quad(4)$$

The weights of the extracted opinion expressions are defined in case 1 and case 2, and the weight for a review is defined in definition 3.2.3. Fuzzy logic is used in the calculation process to make sure the obtained weights are accurate. In order to deeply answer the necessary information of an opinion, the opinion words and the features should be accurately extracted. In the next section, the algorithms of opinion words and feature extraction will be given and the dependency structure will be employed to express the relations between opinion expressions and features.

## 4. Jointly execute opinion mining extraction tasks

Identification and recognition of dependency relations can help to understand the inner associative rules of opinion words and features. Section 4.1 gives a general view of the topology structure of all possible dependency relations to provide the primary elements of opinion mining extraction. Section 4.2 defines all possible working rules between opinion words and features. The opinion mining algorithm is proposed in section 4.3 based on the analysis of section 4.1 and 4.2.

### 4.1. Dependency Relations Classification

Tree structure is quite common to be used to represent syntactic structure of linguistics (Cer et al., 2010; Wu et al., 2009; Hou et al., 2011). Trees are used for reflecting a constituency structure and a dependency structure. Projective structure means there is no edges that cross with other edges. A non-projective tree structure is used to deal with more complex sentences (McDonald et al., 2005; Bohnet and Nivre, 2012; Martins et al., 2010). For constituency structure of the tree, the nodes reflect smaller or larger segments (POS tagger) of the input string, and the edges represent the composition of larger phrases from elements (words). The dependency structure regards the 'verb' as the root of all possible phrase structures. All other syntactic words are either directly or indirectly dependent on the verb. We adopt the dependency structure because of the effectiveness to analyze and parse the information.

The dependency relations refer to syntactic properties and semantic properties which are represented by different layers of dependency structures. Dependency relations between word A and B (A→dep→B) mean A governs B or B depends on A. A dependency relation is considered as double direction and shown as A↔dep↔B in this paper, which means once word A and B having dependency relation *dep*, then searching tasks can be built in two directions. The dependency relations are subject (subj), object (obj), conjunct (conj), modifier (mod), etc. Two categories involving explicit and implicit dependency relations are defined to summarize all possible dependencies to be a new kernel in sentences. The categories defined in our paper are different with Qiu et al. (2011), because our work focuses on discovering the relations on the phrase level while Qiu et al. (2011) on the word level.

**Category 1 (Explicit Dependency Relation (EDR))** The EDR indicates that word A can be found from word B through dependency relation without or with only one additional word from each dependency path in one direction.

**Category 2 (Implicit Dependency Relation (IDR))** The implicit dependency relations indicate that A depends on B through more than one additional word or they both depend on the connection word (H) through more than one additional word.

In figure 2, H1 called middle word, which helps to build the relations between words, and H is called connection word, which connects two relations. In figure 2.1, A depends on B without additional words or with only one middle word in one direction. The two situations are regarded as a basic unit to build a new kernel representing a new restructured phrase unit. In figure 2.2, A and B directly depends on H respectively through only one middle word (H1), then A depends on B directly. In this paper, the number of the middle words varies from zero to one. The number of the connection word only equals to one. The relations of A and B belong to a direct dependency relation (EDR) when the number of middle word equals to zero in Figure 2.1 and Figure 2.2; the relations of A and B belong to an indirect dependency relation (EDR) when the number of middle word equals to one; otherwise, the relations of A and B belong to IDR.

IDR denotes all of the dependency relations apart from EDR. Moreover, IDR is significantly more difficult to discover and tends to incur more errors with ambiguity or fuzzy text. The IDR extraction task can be finished by several EDR subtasks. For example, the sentence: *The camera takes great pictures in low and artificial light. I highly recommend this camera for this reason.* The dependency relations of this sentence are shown in Figure 4. The complete dependency relations can be extracted by extracting six EDRs. We find that the feature *picture* depends on the opinion word *great* through relation *amod*. We also get another feature *light* through picture, and then collect the adjectives that describe the situation of the light; the feature *picture* can also find the potential opinion word *recommend* through middle word *takes*, and the opinion degree *highly* can be obtained through the opinion word *recommend*. The final extracted information of this sentence is shown in Table 6.

Table 6. Examples of the extracted information from unstructured text

| Product feature($f_{ij}$) | Opinion ($oo_{ijkl}$) | Final relations ($r_{ijkl}$) |
|---|---|---|
| \<camera\>;\< pictures \>; \<light\>;\<camera\>; \<reason\> | \<great\>; \<low\>; \<artificial\>; \<highly\>; \< recommend \> | \<camera great pictures low artificial light\>; |
| | | \<highly recommend camera\> |
| Dependency relations: great→amod→pictures; pictures ←prep←in←pobj←light; low→amod→light; light←amod←low; low←conj←artificial; pictures→dobj→takes→dep→recommend; recommend←advmod←highly | | |
| First level Kernels: (great pictures); (low artificial light); (highly recommend) Second level kernels: (camera; great pictures; low artificial light); (highly recommend; camera) | | |

The EDR and the IDR have two common properties (symmetric and transitive) based on their definition and topology structure. The two properties can help to find the propagated directions of the dependency relations, which can also improve the extraction recall and precise degree as well as retrieve the target more quickly.

**Property 1**: *Symmetric dependency relation*: A→dep→B==B←dep←A. The property means the relation between A and B is built and can be searched in both sides once one side relation is built. For example, in Figure 2.1, from word A can find word B based on dependency relation and from word B can also search word A once one side relation is built.

**Property 2**: *Transitive dependency relation*: A→dep→H1→dep→B==A→dep→B. The property means the relation can be transferred from a middle word or several middle words. One constraint for this property is that the transition direction must be the same for all words. For example, it has the relation: *good→amod→zoom→conj→camera*, then we can deduce *good→dep→camera*.

The two properties can help to find the correct dependency relations kernel. The kernel is the combination of two points which have the shortest distance. The second layer kernel is built based on the shortest distance of the first layer kernel. First layer kernel means the dependency relations in word level; second layer kernel means the dependency relations in phrase level.

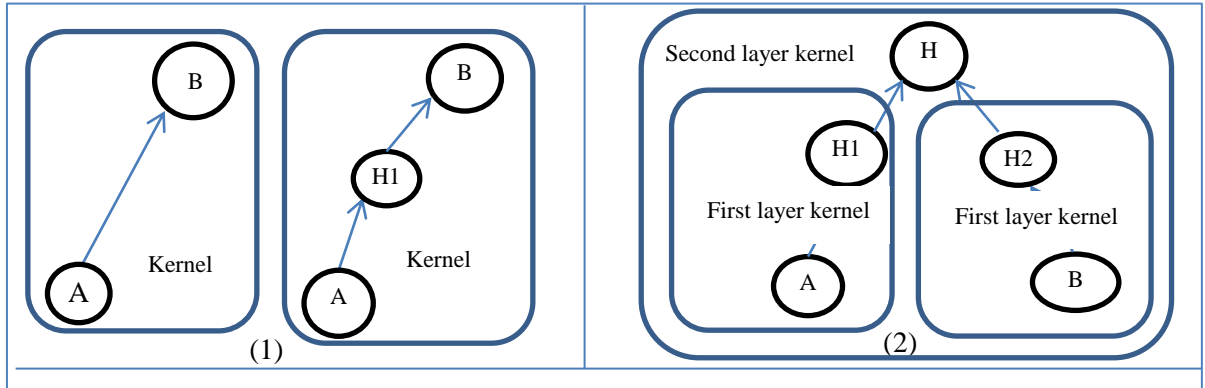

Figure 2. Topology structure of dependency relations

### 4.2. Extraction rules defined based on dependency relations

The extraction is mainly between features and opinion words. For convenience, some symbols are defined as easy reusability. The relations: between opinions and features are defined as FO↔Rel, between opinion words themselves are OO↔Rel, and between features are FF↔Rel. Four basic extraction tasks are defined to separate information extraction: (1). Extracting products' features by using opinion words (FO↔Rel); (2). Retrieving opinions by using the obtained features (OF↔Rel); (3). Extracting features by using the extracted features (FF-Rel); (4). Retrieving opinions based on the known opinion words (OO-Rel). The rules of these four tasks are defined mainly for EDR. Two more tasks are defined by extending these four rules to complete extractions: (5). Extracting products' features by using both the extracted opinion words and the related features; (6). Retrieving opinions based on the extracted opinions and features. The two more added tasks focus on IDR, which are especially used for long distance dependence. Six categories of running rules are clarified and depicted in Table 7.

In Table 7, o (or f) represents the obtained opinion expressions (or features). O (or F) is the set of given or obtained known opinions (or features). POS (O/F) means the POS information that contains the linguistic category of words, such as *noun* and *verb*.{NN, NNS, JJ, RB,VB} are POS tags. O-Dep, that represents opinion word O, depends on the second word based on O-dep relation. F-dep means the feature word F depends on the second word through

F-dep relation. MR={nn, nsubj, mod, prep, obj, conj, dep}, 'mod' contains {amod, advmod}, 'obj' contains {pobj, dobj}. Finally, rules ($R1_i - R6_i$) are formalized and employed to extract features (f) or opinion words (O) based on the previously defined six tasks.

Table7. Rules for features and opinion expressions extraction

| Rule | Input | Representation Formula | Output | Example |
|------|-------|------------------------|--------|---------|
| $R1_1$ (EDR) | O | $O \xrightarrow{\text{Depend (O-Dep)}} F;$ where, $O \in \{O\}$, O-Dep $\in \{MR\}$, POS(F) $\in \{NN, NNS\}$ | f=F; FO↔Rel | Canon PowerShot SX510 takes **good** *photos*. (good←amod→photos) (Figure 3) The **images** are **excellent**. (excellent←nsubj←images) |
| $R1_2$ (EDR) | O | $O \xrightarrow{\text{O-Dep}} H \xrightarrow{\text{F-Dep}} F$ s.t. $O \in \{O\}$, O / F-Dep $\in \{MR\}$ POS(F) $\in \{NN, NNS\}$ | f=F FO↔Rel | The Canon PowerShot SX510 *HS* is a very **good** *value* thanks to a new sensor. (good→amod→value←nsubj←HS) |
| $R1_3$ (EDR) | O | $O \xrightarrow{\text{O-Dep}} H \xrightarrow{\text{F-Dep}} F$ s.t. $O \in \{O\}$, O / F-Dep $\in \{MR\}$, POS(F) $\in \{NN, NNS\}$ | f=F FO↔Rel | It works **great** for a kindle camera. (great←prep←for←pobj←camera) |
| $R2_1$ (EDR) | F | $O \xrightarrow{\text{O-Dep}} F;$ s.t. $F \in \{F\}$, POS(O) $\in \{JJ, RB, VB\}$ | o=O OF↔Rel | Same as $R1_1$, *photos* as the known word and *good* as the extracted word. |
| $R2_2$ (EDR) | F | $O \xrightarrow{\text{O-Dep}} H \xrightarrow{\text{F-Dep}} F$ s.t. $f \in \{F\}$, O / F-Dep $\in \{MR\}$ POS(O) $\in \{JJ, RB, VB\}$ | o=O OF↔Rel | Same as $R1_2$, *HS* as the known word and *good* as the extracted word, also extract the middle word *value* |
| $R2_3$ (EDR) | F | $O \xrightarrow{\text{O-Dep}} H \xrightarrow{\text{F-Dep}} F$ s.t. $f \in \{F\}$, O / F-Dep $\in \{MR\}$ POS(F) $\in \{JJ, RB, VB\}$ | o=O OF↔Rel | Same as $R1_3$, *camera* as the known word and *great* as the extracted word. (camera→pobj→for→prep→great) |
| $R3_1$ (EDR) | F | $F_{i(j)} \xrightarrow{F_{i(j)}\text{-Dep}} F_{j(i)}$ s.t. $F_{i(j)} \in \{F\}$, $F_{i(j)}$-Dep $\in \{conj\}$ POS$(F_{i(j)}) \in \{NN, NNS\}$ | f=F FF↔Rel | It takes breathtaking **photos** and great **videos** too. (photos→conj→videos) |
| $R3_2$ (EDR) | F | $F_{i(j)} \xrightarrow{F_{i(j)}\text{-Dep}} F_{j(i)}$ s.t. $F_{j(i)} \in \{F\}$, $F_{i(j)}$-Dep $\in \{NN\}$ POS$(F_{i(j)}) \in \{NN, NNS\}$ | f=F FF↔Rel | The image **quality** is great. quality←nn←image |
| $R3_3$ (IDR) | F | $F_i \xrightarrow{F_i\text{-Dep}} H \xrightarrow{F_j\text{-Dep}} F_j$ s.t. $F_i \in \{F\}$, $F_i$ / $F_j$ – Dep $\in \{MR\}$ POS$(F_j) \in \{NN, NNS\}$ | f=F FF↔Rel | **SX500** has a smaller camera and a good sized zoom. (SX500←nsubj→has←dobj←camera←conj←zoom) Canon PowerShot *SX510* takes significantly better indoor ***photos.*** (photos→dobj→takes←nsubj←SX510) |
| $R4_1$ (EDR) | O | $O_{i(j)} \xrightarrow{O_{i(j)}\text{-Dep}} O_{j(i)},$ s.t. $O_{j(i)} \in \{O\}$, $O_{i(j)}$ – Dep $\in \{advmod, conj\}$, POS$(O_{i(j)}) \in \{RB\}$ | o=O OO↔Rel | Canon PowerShot *SX510* takes significantly **better** indoor *photos*. (better←advmod←significantly) This camera is **light** and easy to hold. (light←conj←easy) |
| R4$_2$ (EDR) | O | $O_i \xrightarrow{O_i-Dep} H \xleftarrow{O_j-Dep} O_j$, s.t. $O_i \in \{O\}$, $O_i - Dep == O_j - Dep$, $POS(O_{i(j)}) \in \{JJ\}$ | o=O OO↔Rel | If anybody wants a new light, smart, easy use camera, I highly recommend Canon PowerShot. (new→amod→camera←amod←light; new→amod→camera←amod←smart;…) |
| R5$_1$ (IDR) | O | $R1_1 + R3_2$ | f=F FO↔Rel | Canon PowerShot *SX510* takes significantly **better** indoor *photos*. (**better**→amod→**photos**) (**photos**→dobj→takes←nsubj←**SX510**) |
| R6$_1$ (IDR) | O | $R1_1 + R2_3 + R1_1$ | o=O OO↔Rel | The camera takes great pictures in low and artificial light. I highly recommend this camera for this reason. (great→picture→dobj→takes→dep→recommend←highly) (Figure 4) |

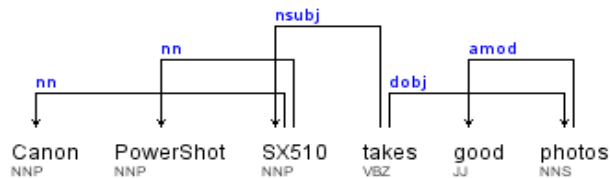

Figure 3. The dependency structure for the sentence: *Canon PowerShot SX510 takes good photos*

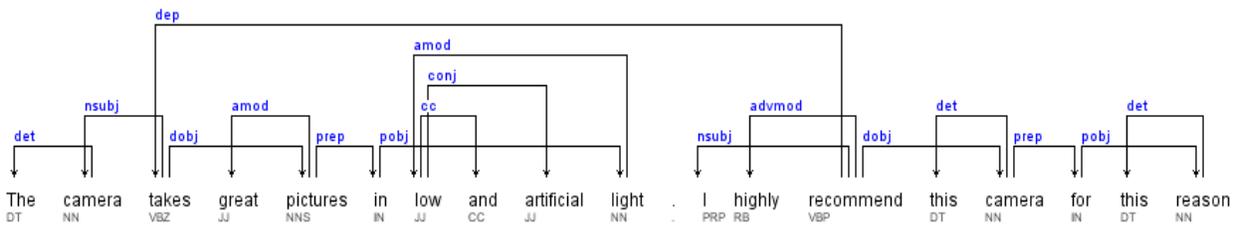

Figure 4. The dependency structure of the sentence: *The camera takes great pictures in low and artificial light. I highly recommend this camera for this reason.*

**Remark 5:** *Details explain about the rules in Table 7.* Rule R1$_i$ is employed to extract features (f) and the potential opinions based on the extracted words (O); R2$_i$ is used to extract opinions (o) based on extracted features (F); R3$_i$ is adopted to obtain the new features (f) based on extracted features (F). R4$_i$ is given to gain the opinions (o) based on the existing words (O). R5$_i$ is described to give the products' features (f) by using known opinion words (O) and the related features. R6$_i$ is used to retrieve the opinions based on the extracted opinions and features. Normally, several rules are combined to finish one extraction. For example, we have the first example "Canon PowerShot SX510 takes **good** *photos*", which the related dependency tree is depicted in Figure 3. We obtain *good* as an opinion and this word relies on *photos* through *amod*. '*Photos*' is labeled as NNS, which is a feature based on the rule R1$_1$. If the output is only *(good photos),* some features are lost. Therefore, we continue to find the closest features *SX510* based on the rule R1$_2$ and R5$_1$ through the feature *photos*.

### 4.3. Opinion mining extraction algorithm

Table 8 shows the detailed opinion mining extraction algorithm. The initial values of the proposed algorithm are shown as: opinions dictionary O, the opinion degree intensifiers OD, and the review data RD. This algorithm adopts a single review from customers as the basic analysis unit. For each review, anytime the customer mentions a feature name, such as camera, those words are considered unique and should be excluded from the analysis. In other words, if the review talks about the "camera's zoom" feature, and afterwards the same word "zoom" appears again in the same review; the word "zoom" will be excluded from being analyzed further. This assumption determines the stop point of the proposed algorithm. If no new feature words are found in the review, then the algorithm will stop its analysis for the current review and begin to analyze the next review.

## Table 8a Algorithm 1: opinion mining extraction algorithm

**Algorithm Opinion_Mining_Extraction()**

**Input:** Opinion word dictionary O, Opinion Degree Intensifiers OD, Review Data:RD
**Output:** The set of features F, the set of expanded opinion words EO, the opinion polarity (or orientation) for a product: OW
**BEGIN**

1. Expanded opinion words: EO=$\varnothing$ ; F=$\varnothing$ ; ODI=$\varnothing$
2. **For** each dependency parsed review $RD_k$
3. **// Obtaining the initial opinion words and intensifier degree words in** $RD_k$ **based on the dictionaries of** O **and** OD
4.    **for** each word tagged JJ,RB, and VB in $RD_k$ [2]
5.      Traversing the $RD_k$, and extracting the opinion words ($OP_i$) if they are appearing in O; i++;
6.      Extracting new opinion words {$OP_j$} in $RD_k$ by using the Rules $R4_1$-$R4_2$ based on extracted opinion words {$OP_i$}; j++;
7.      Inputting the obtained $OP_i$ and $OP_j$ into EO, and then EO={$OP_{[1,...,i]}$ , $OP_{[1,...,j]}$ }(for short EO={ $OP_{1-i}$, $OP_{1-j}$ });
8.      Traversing the $RD_k$, and extracting the degree intensifier words ($DW_d$) if they are appearing in OD;
9.      Inputting the obtained $DW_d$ into ODI, and then ODI={$DW_{1-d}$}; d++;
10.    **End for**
11. **//Extracting the features based on the obtained initial opinion words and opinion degree intensifier words**
12. Extracting features {$F_{fi}$} in $RD_k$ by using the Rules $R1_1$-$R1_3$ based on opinion words EO={$OP_{1-i}$, $OP_{1-j}$ }; fi++;
13. *if* (Extracted new features not in F)
14.    Extracting features {$F_{fj}$} using Rules $R3_1$-$R3_3$ based on the new extracted features {$F_{fi}$}; fj++;
15.    *Extracting and updating new opinion words {$OP_{1-p}$} using Rules $R2_1$-$R2_3$ based on extracted features F={$F_{fi}$, $F_{fj}$ };*
16.    *Extracting new features {$F_{fp}$} in $RD_k$ by using the Rules $R1_1$-$R1_3$ based on new opinion words EO={$OP_{1-p}$}; fp++;*
17. **End if**
18. Setting F={$F_{fi}$, $F_{fj}$ , $F_{fp}$ }; EO={$OP_{1-i}$, $OP_{1-j}$, $OP_{1-p}$ };
19. KernelFeature_OpinionSets=Build_kernel(F, EO, $RD_k$);
20. Recording appearing frequency af of EO based on related F;
21. *if* the opinion words EO have the corresponding degree intensifier ODI
22.    Building triple {ODI, EO, F}
23.    ***Else if***
24.    *Building triple {null, EO, F}*
25.    ***End if***
26. Unique and update {ODI,EO,F};
27. Calculating the opinion polarity{OW} based on **Definition 3.2.1- 3.2.3, Triple {ODI, EO, F}, and af;**
28. **End for**
**END**

## Table 8b Algorithm 2: Building kernel algorithm

**Algorithm [KernelFeature_OpinionSets]=Build_kernel(F, EO, $RD_k$)**

**Input:** obtained features F, obtained opinions EO, Review Data ($RD_k$)
**Output:** dependency relations in word level, dependency relations in phrase level
**BEGIN**

1. Record the distance between words based on dependency relations;
2. Build the first layer kernel based on the obtained features F and opinion words EO by selecting the shortest distance;
3. **for** every sentence of the $RD_k$
4.    Build the second layer kernel based on the first layer while considering the opinion phrases pattern in Table 1;
   // the algorithm only build the kernel till to the second layer by considering the running time parameter;
5. **End for**
**END**

---

[2] For the opinion words, we traverse the words tagged with JJ, RB, and VB. For the degree words, we traverse the whole review.

**Table 9** An illustration to demonstrate the working of Opinion mining extraction algorithm

| Sequence | Original words | Extracted words | Rule | Specific Line in algorithm | Triple |
|---|---|---|---|---|---|
| 1 OO↔Rel | nice | easy | $R4_1$ | Line 6 | (null, nice,easy) |
| 1 OF↔Rel | great | quality | $R1_1$ | Line 12 | (null, great,quality) |
| 2 OF↔Rel | Love | Features | $R1_3$ | Line 12 | (null, love, features) |
| 3 OF↔Rel | neat | | | Line 12 | (very, neat, null) |
| 4 OF↔Rel | recommend | button | $R1_3$ | Line 12 | (highly, recommend, button) |
| 5 OF↔Rel | nice | button | $R1_3$ | Line 12 | (null, nice, button) |
| 6 OF↔Rel | easy | button | $R1_3$ | Line 12 | (null,easy,button) |
| 7 OF↔Rel | love | | | Line 12 | (very much, love, null) |
| 8 OF↔Rel | Excellent | images | $R1_1$ | Line 12 | (so, excellent, images) |
| 1 FF↔Rel | quality | image | $R3_2$ | Line 14 | (null, great, <image, quality>) |
| 2 FF↔Rel | problem | delay | $R3_3$ | Line 14 | (n't, null,<problem, delay>) |
| 3 FF↔Rel | button | camera | $R3_2$ | Line 14 | (null, easy ,<camera, bag>) |
| 4 FF↔Rel | images | videos | $R3_1$ | Line 14 | (so, excellent, <images, videos>) |
| 1 FO↔Rel | problem | shooting | $R2_3$ | Line 15 | (n't, shooting,<problem, delay>) |
| 2 FO↔Rel | Features | gimmicky | $R2_1$ | Line 15 | (null, gimmicky, features) |
| 3 FO↔Rel | button | easy | $R2_1$ | Line 15 | (null, easy, button) |
| Feature-opinion relations (line 26) | (null, great, <image, quality>); (null, <love, gimmicky>, features); (very, neat, null); (n't, shooting,<problem, delay>); (highly, <recommend, nice, easy>, <camera, button>); (very much, love, null); (so, excellent, <images, videos>) | | | | |

An example is proposed to demonstrate the running mechanism of the proposed algorithm. Suppose the following review for Canon PowerShot A2400: *The image quality is great. I love all the gimmicky features, some of which I used to create with Photoshop later, thus it saves me time. It looks very neat and until now I haven't had any problem with the shooting delay I read about. I highly recommend to use this nice easy camera button. Overall I love it very much and the images and videos are so excellent!* Firstly, we traverse and compare with the opinion dictionary; the opinion words are extracted, such as: ***great, love, neat, recommend, nice, love, excellent, and problem***. *The degree words are also extracted, such as: **very, highly, very much, n't, and so***. These obtained words are used as the initial input of the algorithm. The running steps for the proposed example are shown in Table 9. Table 9 shows that the product features and new opinions in the review are perceived without manual intervention, which means the proposed algorithm is an unsupervised learning algorithm.

**Remark 6:** This algorithm directly groups two words into one kernel when the two words are in an equivalent structure. The equivalent structure includes the dependencies of *nn* and *conj*. In addition, two words which share a common father in the dependency path are also equivalent structure. For example, in Table 9, ***<u>images</u> and <u>videos</u>*** *are so excellent*, group *images, videos* into one kernel <images, videos>; Similarly, we have <image, quality>, <love, gimmicky>, and etc.

## 5. Experiment Results and Discussions

Previous works for opinion mining can be divided into two directions: sentiment classification and information extraction. The former direction aims to identify positive and negative sentiments from a text. Sophisticated sentiment classification techniques include Naive Bayes (Pappas and Kotsiantis, 2013; Barber, 2012) and SVM (Support vector machine)(Fan et al., 2008). In order to produce richer information, the latter focuses on extracting the elements that can finalize sentiment tasks(Qiu et al., 2011; Jakob and Gurevych, 2010). The elements include opinion words that express an opinion and the features that represent opinion targets. The proposed algorithm has the capability to cover the two directions.

Deep analysis is conducted by comparing with several previous works to evaluate the performance of our method in five aspects: sentiment classification, orientation evaluation and prediction, feature extraction, opinion extraction, and feature-opinion extraction. Precision, recall, F-score, and running times are the parameters that are adopted to judge the effectiveness of the proposed Algorithm. Precision (P) is calculated as part of retrieved instances that are correct. Recall (R) refers to the part of relevant instances that are retrieved. F-score (F) is the harmonic mean of precision and recall.

The raw customer opinion data was collected by using publicly available information from the Amazon site. The experiments were conducted in three domains that include: Canon camera, Casio watch, and Nike shoes. The test data included 3,458 customer reviews of 17 different types of canon cameras, 354 customer reviews of Casio G-Shock watch, and 252 customer reviews of Nike women's shoes.

## 5.1 Performance of sentiment analysis

To increase the credibility of the results, we randomly selected 200 customers' comments for three different products in each test. 10-fold cross validation is conducted for each dataset in each time test, and the results for each parameter are obtained from the average value of the ten times trials. In the proposed algorithm, the weights of the opinion words that are in the dictionary have already been defined, but the new extracted opinion words that are not in the dictionary have not been defined yet. Therefore, in the cross validation, the training sets are used to set the appropriate degree of the new opinion words, and this information will be used in the testing sets. The reviewers have graded the products while giving their comments. We take the grades given by the reviewers as the desired classifications. Then, P, R, and F-score of Naive Bayes, SVM, and the proposed algorithms can be achieved. The detailed results of performance evaluation are shown in Figure 5.

From Figure 5, we can see that SVM performs the worst of all the cases, which means SVM is weaker in determining reviewer's classification. The Naive Bayes algorithm achieves better performance than SVM in four parameters. Our method clearly improved the F-score in comparison with the other algorithms. The recall value obtained from our method significantly outperformed (nearly 15% improvement) the other algorithms on all datasets. The running time of our method is acceptable in all datasets, which is only less than a second slower (we only select the classification time to compare. The reading time of files is not added.). This indicates that our algorithm is effective in defining the opinion polarities.

Naive Bayes and SVM algorithms are confined to predict the positive or negative category of any given document. These algorithms cannot produce the intensity of a positive or a negative and cannot locate the corresponding features with the specifications based on each reviewer's opinion. Comparing consumer opinions of its products and those of its similar subtype cameras to find the products' strengths and weaknesses is crucial for marketing intelligence and for product benchmarking. In order to produce more useful information for the customers and the designers, we use our method to analyze the orientation of different subtype products.

Main articles of the Canon camera include Powershot and EOS. Powershot cameras have 35 different types of products, which is classed into five categories. One category labeled 'High-End, Advanced Digital Cameras' has 15 heterogeneous products. Seven popular selling products are picked out to evaluate the products' performance based on customers' opinions. The 'orientation value' is obtained based on the defined constraints (3.2.1-3.2.3). In order to fully understand the products overall situation, we also collect and depict the price of the products and the total number of reviews. The normalization values of three-dimensional variables in terms of "orientation value", "Product Price", and "Total Number of Current Reviews" are shown in Figure 6. PowerShot S110, SX510 HS, and SX280 HS are the most popular products, which has more than 280 customer reviews on Amazon for each product, whereas SX510 HS is the most favorable product, because its orientation value is nearly 10% higher than SX280 HS and S110. Further studies need to be done to figure out the specific weak points of SX280 HS and S110.

## 5.2. Information extraction analysis on datasets

### 5.2.1 Performance comparison between baseline approaches and proposed method

#### 5.2.1.1 Information extraction performance comparison

The similar products S110, SX510 HS, and SX280 HS have 232, 381, and 517 reviewers respectively. The total number of sentences for each dataset is marked in Table 10. For each sentence of each review, it has five rows that include the sentence itself, POS, dependency relations, detailed dependency relations, and the sequence markers. The sequence markers are F, O, D, and N in the data sets. F denotes the features, O denotes the opinion words, DO denotes the opinion degree intensifier words, and N denotes none of them. The extracted correct number of features (or opinion words) by the proposed algorithm can be obtained by comparing with the sequence markers. The total number of features (or opinion words) can also be calculated based on the sequence markers. Therefore, the P, R, and F-score in Table 10 can finally be achieved. We generate the experiments results in sentiment classification, feature and opinion extraction, to make a deeper analysis of the algorithms performance.

As for classification analysis, the same observations can be made with Figure 5. The proposed method can better recognize and assign more accurate orientation values. Figure 5 and 'classification' dimension in Table 10 (rows 1-4, 11-14, and 21-24) demonstrate that the proposed method is more effective than the other algorithms. The reason for this is that we clearly defined each opinion words' fuzzy scale, considering some adverbs and verbs as the opinion words, and finding the modifier that could give the additional intensity information of an opinion word.

In order to test the information extraction performance, we compare the proposed method with Qiu et al. (hereinafter called Qiu2011), and **c**onditional **r**andom **fi**elds (CRF) (Jakob and Gurevych, 2010, hereinafter called Jakob2010).Qiu2011 adopted dependency parser to identify syntactic relations between opinion words and features, and proposed a double propagation algorithm to do information extraction. Qiu2011 claimed that the proposed propagation algorithm outperforms CRF (Lafferty et al., 2001), Popescu (Popescu and Etzioni, 2007), and Kanayama (Kanayama and Nasukawa, 2006). Jakob2010 argued that the advanced CRF-based algorithm clearly outperforms the baseline algorithms on all datasets, which improves the performance based on F-score in four single domains. Hence, we employ Qiu2011 and Jakob2010 approach as the baseline.

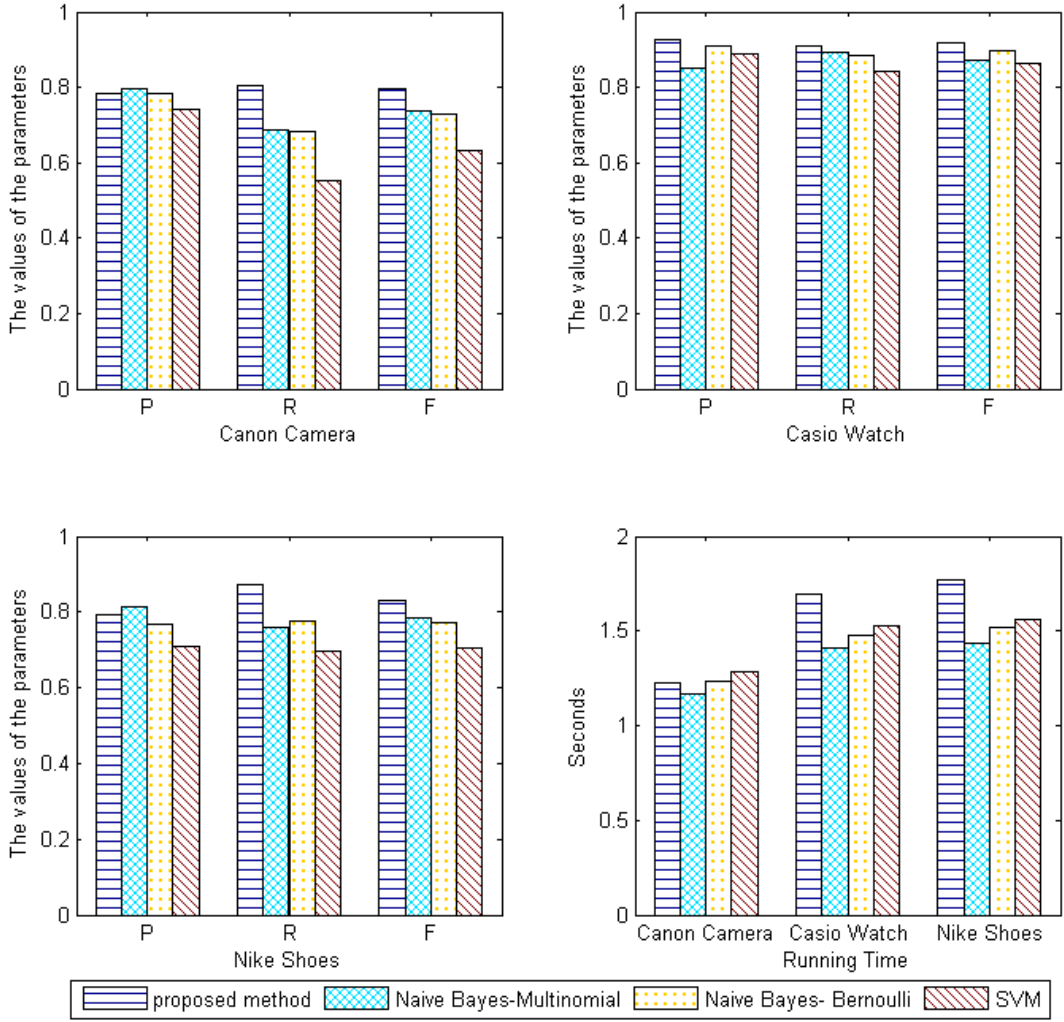

**Figure 5.** Performance comparison of four methods for sentiment classification

Qiu2011 implemented propagation and non-propagation approach for opinion word extraction. The performance showed that the propagation approach achieved the highest F-score. We observe that the newly obtained opinion words have strong relevance with the defined opinion word dictionary. Using the full range of the opinion dictionary we can gain more adequate and effective opinion words. Therefore, the Qiu2011 propagation approach with full range dictionary opinion words is employed. The performance of Qiu2011-propagation approach implemented in our work is higher than the original one; the reason is that we consider some adverbs and verbs as opinion words.

Jakob2010 defined five types of input features for the CRF-based approach. The first type (**Token**, or **TK for short**) input is the string of the current token; the second type (**POS**) represents the part-of-speech tag of the current token; the third type (Short Dependency Path) labels all tokens that have a direct dependency relation to an opinion expression in a single sentence; **wrdDist** (Word Distance) is the closest labeled distance among noun phrases; The type of Opinion Sentence aims to enable the CRF algorithm distinguishing whether a certain token in a sentence contains an opinion or not. Jakob2011 observed that all features in combination obtained the best performance regarding F-score on all datasets. Therefore, we test all features for S110, SX510 HS, and SX280 HS datasets.

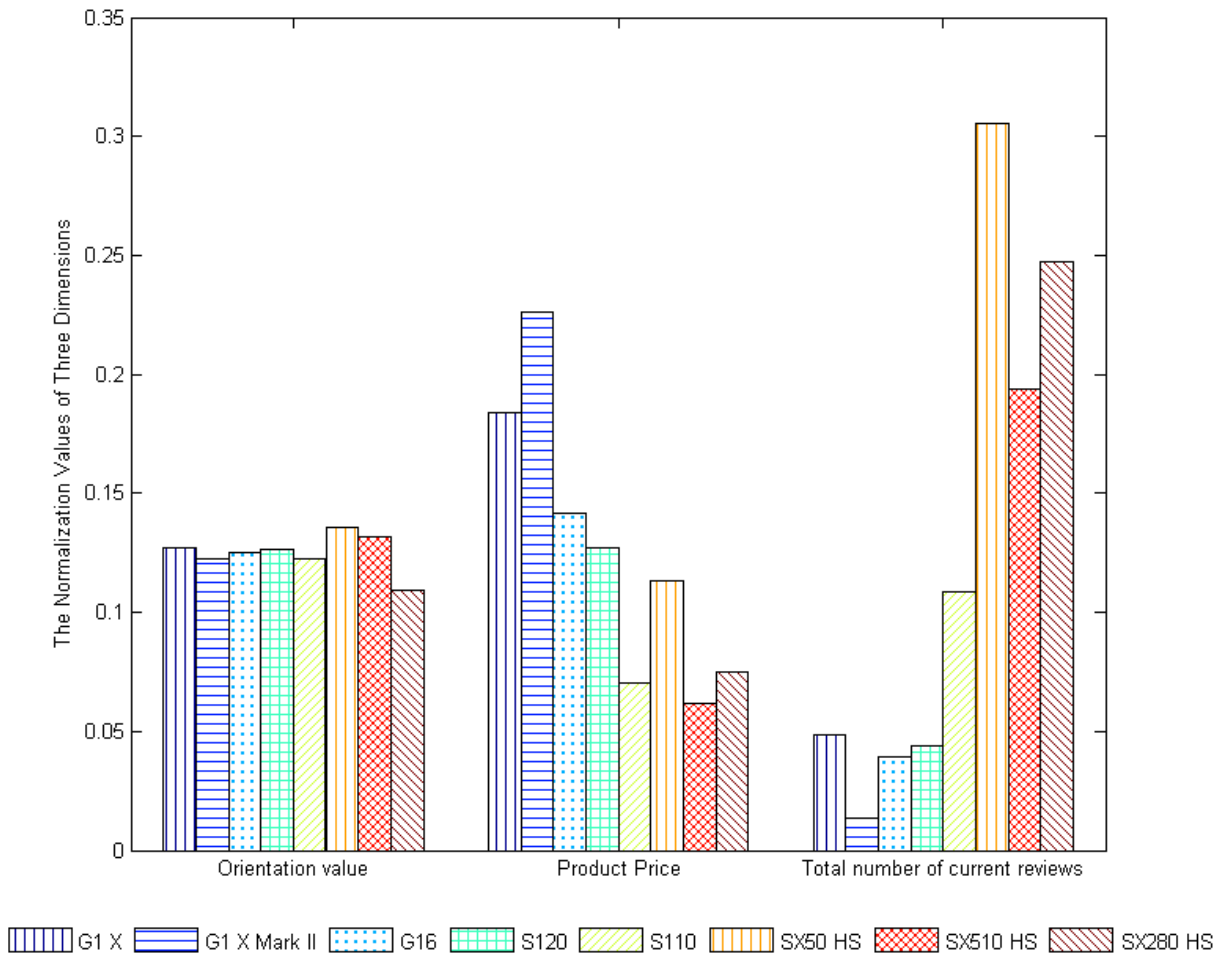

**Figure 6.** Orientation analysis for High-End, Advanced Digital Canon Cameras

Rows 5-7, 15-17, and 25-27 in Table 10 gives the comparison results of different approaches. The precision of feature extraction of our method is 6.43% higher on average than Jakob2011 and Qiu2011 respectively, which means that our method can extract more effective instances among feature elements. The recall of feature extraction is also significantly improved by up to 29.46% and 14.86% on average when comparing with Jakob2011 and Qiu2011 respectively. We observe that our method outperforms better than the other methods for opinion extraction in terms of precision and recall. Meanwhile, the gain in F-score is between 0.6713 in S110 (feature extraction) and 0.8211 in SX510 HS (opinion extraction). The achieved F-score is also higher than the other methods in all datasets. The reason for the superior results is that we match the reviewed data with an intensive opinion words dictionary and consider the dependency relations up to the phrase level by building kernels between closely related words of each sentence. Although the proposed method clearly outperforms the other baseline approaches, the same generation's trend also exists in the individual results: Opinion extraction yields better results than feature extraction. It is because the feature words are more complex and changeable. The opinion words and the obtained feature words are used as guide words to iteratively find new features words, whereas the reviewers may adopt synonyms or analogies to describe the same feature. In general, the comprehensive analysis shows that our method is more effective and more suitable to be used in real-life cases.

**Table 10** Precision, Recall, and F-score of our method, Qiu2011, and Jakob2011

| Selected Product ID (High-End, Advanced Digital Canon Camera) | Directions | Methods | P | R | F |
|---|---|---|---|---|---|
| Canon PowerShot S110<br><br>No. of Reviews: 232<br>Sentences: 2054 | Classification | Our method | **0.9851** | 0.9041 | **0.9429** |
| | | Naive Bayes-Multinomial | 0.9849 | 0.9155 | 0.9489 |
| | | Naive Bayes- Bernoulli | 0.7736 | 0.9111 | 0.8367 |
| | | SVM | 0.7459 | **0.9556** | 0.8378 |
| | Feature extraction | Our method | **0.6575** | **0.6857** | **0.6713** |
| | | Qiu2011 | 0.6139 | 0.6043 | 0.6091 |
| | | Jakob2010 | 0.5714 | 0.4400 | 0.4972 |
| | Opinion extraction | Our method | 0.7625 | **0.8222** | **0.7912** |
| | | Qiu2011 | **0.7778** | 0.7125 | 0.7437 |
| | | Jakob2010 | 0.6625 | 0.7143 | 0.6874 |
| Canon PowerShot SX510 HS<br><br>No. of Reviews: 381<br>Sentences: 2456 | Classification | Our method | **0.8987** | **0.8452** | **0.8712** |
| | | Naive Bayes-Multinomial | 0.8462 | 0.7857 | 0.8148 |
| | | Naive Bayes- Bernoulli | 0.8667 | 0.8125 | 0.8387 |
| | | SVM | 0.8378 | 0.8158 | 0.8267 |
| | Feature extraction | Our method | **0.8046** | **0.6575** | **0.7237** |
| | | Qiu2011 | 0.7241 | 0.4118 | 0.5250 |
| | | Jakob2010 | 0.5172 | 0.2941 | 0.3750 |
| | Opinion extraction | Our method | **0.7812** | **0.8654** | **0.8211** |
| | | (Qiu et al., 2011) | 0.7677 | 0.5135 | 0.6154 |
| | | CRF | 0.6970 | 0.4662 | 0.5587 |
| Canon PowerShot SX280 HS<br><br>No. of Reviews: 517<br>Sentences: 4992 | Classification | Our method | **0.7536** | **0.6000** | **0.6585** |
| | | Naive Bayes-Multinomial | 0.6667 | 0.4615 | 0.5455 |
| | | Naive Bayes- Bernoulli | 0.4000 | 0.6667 | 0.5000 |
| | | SVM | 0.5000 | 0.4737 | 0.4865 |
| | Feature extraction | Our method | **0.6892** | **0.7183** | **0.7034** |
| | | Qiu2011 | 0.6204 | 0.5986 | 0.6093 |
| | | Jakob2010 | 0.4599 | 0.4437 | 0.4516 |
| | Opinion extraction | Our method | **0.7958** | **0.7434** | **0.7687** |
| | | Qiu2011 | 0.6069 | 0.5789 | 0.5926 |
| | | Jakob2010 | 0.4437 | 0.4145 | 0.4286 |

## 5.2.1.2 Extracted opinion words and calculated priorities comparison

This work can also provide a list of opinion words. The priority value of each view is determined by the fuzzy operation of all the obtained opinion words in a review. Rill et al. (2012) (hereinafter called Rill2012) proposed an approach to generate lists of opinion bearing phrases with their opinion values. We compare our work with Rill2012 by analyzing the extracted opinion words and the calculated opinion values. Some comparison examples are shown in Table 11 and Table 12.

In Table 11, in general, the opinion values obtained by the two approaches can reasonably reflect the opinion degree. Also, the shifters change the opinion values in the right direction. The multiple usage of intensifier words lead to reasonable opinion values as well. It is worth mentioning that the opinion value of the phrase 'not very good' is a positive score in our work, but Rill2012 obtained a negative score for this phrase. In Table 12, the sample data follows the same pattern: "Entity Name_Star Rating_Review Title". The scale of reviews is between 1 ('very bad') and 5('very good'). The scale of 3 represents a neutral assessment. The range of opinion words is between -1 and 1. The proposed algorithm focuses on extracting the opinion words in the review text, while Rill2012 analyzed the opinion words in the review title. We compared the two approaches by using the extracted opinion words and adopting the calculated opinion values to represent the priority of each review. The compared results revealed that the opinion words only extracted from the review title cannot accurately reflect the opinion of the reviewer. Moreover, the review titles of 'Good camera' , 'great images' , and 'Very good point and shoot' have the same star rating. However, the numerical values for priorities obtained by Rill2012 differ greatly. The calculated opinion values by our work have more accurately represented the priority of each review. Filtering strategies have been given in the work of Rill2012 to exclude irony and bipolar opinions, whereas our work can analyze these reviews, because the reviewers tend to give a clear description in the review text. Some verb- and noun-based phrases are also included in this work, while Rill2012 only considered adjective-based phrases.

**Table 11** Some comparison examples of opinion values for phrases based on the adjective "good"

| Phrase | Opinion values (Rill2012) | Opinion values (our work) |
|---|---|---|
| So good | 0.831 | 1.000 |
| Really good | 0.798 | 0.9333 |
| Very good | 0.755 | 0.9333 |
| So far so good | 0.719 | 0.6567 |
| **Not very good** | -0.599 | 0.6567 |
| good | 0.560 | 0.5 |
| Pretty good | 0.442 | 0.2333 |
| Not good | -0.637 | -0.2767 |

**Table 12** Comparisons of two approaches for obtained opinion words and calculated priorities

| Review(Entity Name_Star Rating_Review Title) | Opinion words (proposed algorithm) | Priority (proposed algorithm) | Opinion words (Rill et al., 2012) | Priority (Rill et al., 2012) |
|---|---|---|---|---|
| Canon PowerShot S110_4_GOOD CAMERA | fine; much better; like; quite; easy to use | 0.75 | Good | 0.560 |
| Canon PowerShot S110_4_great images | Amazing; not easiest; high; great; | 0.7167 | Great | 0.846 |
| Canon PowerShot S110_4_Very good point and shoot | significantly better; bothering | 0.7333 | Very good | 0.755 |
| Canon PowerShot SX280 HS_1_Good photos, but HUGE-LY flawed camera do not even consider it | Error; NOT; not; n't; error; unfortunately | -0.8667 | NULL | NULL |

### 5.2.2 Feature-by-feature comparison among heterogeneous products based on the extracted information

We make ***feature-by-feature comparison*** of reviews regarding the extracted information. For the product designer, feature comparisons can accurately figure out the weaker dimensions as well as recommend the good values to improve the weaker dimensions. For a potential customer, feature comparison could allow the buyers to see the opinions from the existing customers. The features are separated into technical and non-technical groups that could help to describe the strengths and the weaknesses of each product. The ***technical features*** include basic fundamental dimensions, such as: image quality, battery, weight, video, zoom, and wifi capability of camera products. **Technical matrix** is used to express the comprehensive evaluation of existing technical features. The ***non-technical features*** have product usability and user satisfaction. **Product usability** describes the quality of user experience across products, which is a synthetic evaluation of 'easy to use', 'easy to operate', and 'user friendly'. **User satisfaction** is used to measure whether users are satisfied, tolerating, or dissatisfied after they consume the product. The dimensions that are used to assess user satisfaction include the 'user's expectations in comparison with perceptual experience', 'Likelihood to recommend to others', and 'intention of repurchase'.

The content of technical and non-technical features is determined by the frequently extracted terms of features and the related opinion words (hereinafter called feature-opinion). The extracted frequent terms of feature-opinion are grouped into the defined feature dimensions based on their semantic similarity. The orientation values of extracted feature-opinion are assigned based on formulas 3.2.1-3.2.3. Figure 7 gives the feature-by-feature comparison results.

We first analyze technical feature-by-feature comparison results in Figure 7. SX280 HS is the weakest product in terms of battery and video dimension. More than half of the obtained feature-opinion in the battery dimension referred to a terrible battery quality, such as: 'bad battery life', 'battery died', 'battery drains', 'disappointed battery', and 'battery indicator issue' (Top 5 extracted negative frequent terms from the proposed algorithm). The positive value of video dimension is obviously lower than same type products. Negative frequency terms of extracted feature-opinion from the proposed algorithm are 'video problem', 'video issues', 'video shuts off', 'video not work', and 'disappointed video performance'. Moreover, 39.84% of extracted terms point to 'short battery life' and 21.48% are obtained as 'battery indicators issues'. Battery indicator issues are mainly about the indicators misleading the actual state of charge of the battery. The results also have 57 terms like 'defective firmware upgrade' in battery dimension. Therefore, we report the poor battery dimension, because of the battery life and the indicator problem; and the proposed solution from the company doesn't completely solve the problem. As for the video dimension, the extracted results are more inconsistent and disorganized, such as: 'video camera died', 'minutes video battery shut(s) off', and 'zoom video mode battery shut down'. We can deduce that the video problem is probably caused by a battery problem. The negative polarity value of image quality, battery, video, and wifi capability for S110 is 6.03%, 9.48%, 3.45%, and 4.74% respectively, which is a small degree in comparison with the SX280 HS. However, the negative polarity is still higher than SX510 HS. From the obtained technical feature-opinion results, we can see that SX510 is more favorable, because it has a high comprehensive performance.

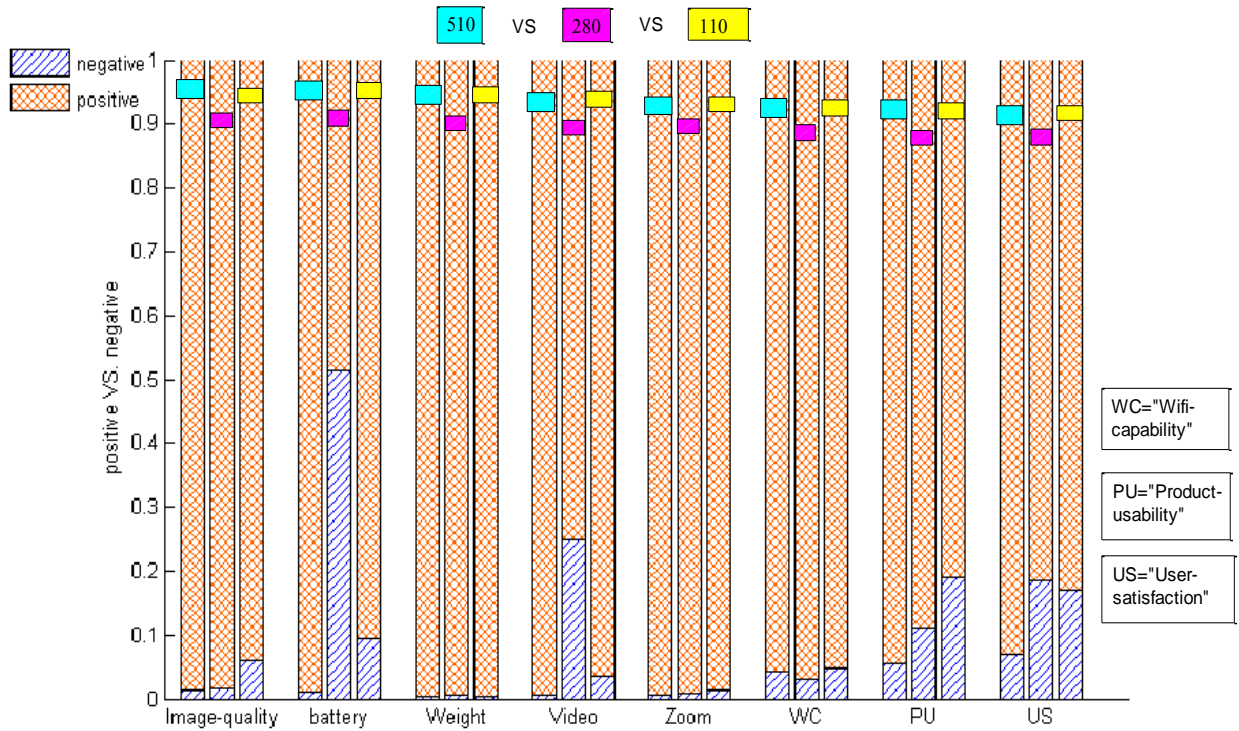

**Figure 7.** Feature-by-feature comparison of three products: SX510 HS ("510"), SX280 HS ("280"), and S110 ("110")

Figure 7 reveals that S110 achieved the poorest performance in non-technical feature dimensions. 19.05% extracted feature-opinion terms in product usability dimension represent that the usability of this product is defective, which is much worse than the other products. S110 and SX280 HS have a lower performance in user satisfaction dimension. Seven obtained terms transmit the information that the corresponded reviewers felt extremely disappointed with the wifi capability of product S110 and recommended customers not to buy the canon camera anymore. Twelve extracted terms show that the correspondent reviewers have a strong disappointed emotion and declare that they will not buy a canon product ever again as a result of the battery issues of SX280 HS.

### 5.2.3 Improvement strategies based on prediction values in linear regression relationships between orientation value, features, and consumption quantity

**O**rientation **v**alue (OV) and **c**onsumption **q**uantity (CQ) are key impact factors to represent the degree of one product's success. The higher the orientation value, the higher probability a customer intends to make a first time purchase. The company will gain more profits with higher CQ. The OV of a specific product is determined by the combination of technical and non-technical features, and CQ has implicit relations with product features. We aim to improve the OV while maintaining the growth trend of CQ. Hence, we go to discover the changing trend of OV and CQ by feature-varying trend changes, and figure out the relations among the polarities of features, orientation values, and consumption quantity.

The orientation analysis in Figure 6 and feature analysis in Figure 7 reveals that SX510 HS is more successful than the same type products. Therefore, we select SX510 HS as the benchmark product to study the changing trend of OV and CQ of the other products. The basic technical and non-technical feature dimensions are considered to have equal contribution to the overall feature score. The technical and non-technical features are calculated separately based on extracted feature-opinion results. The OV is obtained by the defined constraints (3.2.1-3.2.3). The CQ is the number of buyers. In this paper, the consumption quantity cannot be directly obtained from the Amazon website. Once the number of reviews is set, then the consumption quantities are determined. The number of reviews for every month is directly obtained from Amazon. Afterward, the consumption quantity for every month can be determined. To be more clear, the researchers in market domain hold the idea that higher number of reviews is associated with higher sales (Chen et al., 2004; Chevalier and Mayzlin, 2006). Therefore, the consumption quantity is associated in a positive way with the number of reviews every month. This article focuses on the overall trend, so it is not needed to look for the coefficients of correlation between the reviews and consumption quantity, because it will give the same analysis. Therefore, we have adopted the number of reviews as a substitute for consumption quantity.

In order to visualize the general changing trend of SX280 HS, we will use 18 months of data about the technical and non-technical negative polarity value, orientation value, and consumption quantity from May, 2013 to September, 2014 as depicted in the upper part of Figure 8. In order to better display the overall changing trends, the value of technical and non-technical features are enlarged ten times and the CQ value is reduced ten times by comparing with the original obtained value. The general view of obtained data shows that orientation value, technical and non-technical negative polarities have a slight fluctuation around the points of 3.8, 0.1595, and 0.1262 respectively. Moreover, the CQ is more variable and reaches to the absolute maximum value in January, 2014, which means this indictor changes frequently and product SX280 HS has its maximum buyers at this point.

Detail analysis is conducted to answer which dimension(s) should be improved to achieve a better OV while raising the number of consumption quantities. Four dimensions, including: battery, video, product usability, and user satisfaction, are further studied based on the fact revealed in Figure 7. The calculation data is stored into a 6*18 matrix with these four feature dimensions and the related dimensions of OV and CQ. Linear regression is a statistical modeling technique, which is adopted to describe the relationship between the response variable (OV or CQ) as a function of independent variables (feature dimensions, also called predictors). The mathematical relationships are built between the response variable and predictors when the value of root-mean-square error approaches to the lowest point. Root-mean-square error represents the sample standard deviation of the differences between predicted values and observed values. The obtained prediction values are shown in the lower part of Figure 8, meanwhile, the shown values of CQ are reduced ten times to make a better display in figure 8.

The obtained predication results indicate that improving the dimensions of battery and product usability can efficiently improve OV and CQ of SX280 HS. The video dimension has a strong correlation with the battery dimension. Improving the video dimension by a large degree can only slightly increase the predicated OV and CQ. By improving the user satisfaction dimension, OV and CQ will not be increased. Therefore, we recommend improving the dimensions of battery and product usability.

A similar analysis is conducted for product S110. For product S110, increasing the battery and video dimensions can cause a slight improvement of OV, yet it will partly cause a decrease of CQ. Improving wifi capability dimension cannot bring improvement of OV and CQ. The improvement of the image quality dimension can bring an increase of OV and CQ to a small extent. Raising values in the dimensions of product usability and user satisfaction can significantly increase OV and CQ. The obtained results show that product S110 can greatly benefit from increasing non-technical features.

### 5.3 Discussion

### 5.3.1. The impact of opinion words dictionary size on information extraction

Qiu2011 argued that the propagation approach can retrieve opinions and features iteratively by using only a small size of the opinion dictionary. However, we can only get a small number of extracted features and opinion words by using a small sized opinion dictionary in both Qiu2011 and our algorithm. Hence, we study the effects of opinion dictionary size on extracted numbers of opinion words and features. We use the original opinion lexicon to randomly produce 10%, 20%, 50%, 80%, and 100% of the complete opinion lexicon. Two important indicators are needed to investigate. One is the average number of initial extracted opinion words per each review, which is calculated based on the defined size of opinion lexicon. The other one is the average number of new extracted opinion words per each review, which is separately obtained by our algorithm and Qiu2011. We study the same datasets with Table 10 and the results are depicted in Table 13.

The analysis in Table 13 reveals that the size of the opinion dictionary has a strong influence of information extraction. The two methods are hardly getting any extraction content for three datasets when the size is 10%. As the opinion dictionary increases in size, the obtained and extracted opinion words also increase. Both methods gain the maximum number of opinion words when using the full range dictionary. Our method can obtain more opinion words in five different sizes by comparing with Qiu2011.The extracted opinion words contain duplicated words. The intensity words such as *n't, not, very* are also extracted for each review as an intensity dictionary is also proposed. With the increased size of the opinion dictionary, the extracted content is richer. It is possible that the bigger the dictionary, the more redundant information is given. Therefore, we go further to study how many opinion words probably exist and should be extracted in a single review document and then give the proper dictionary size.

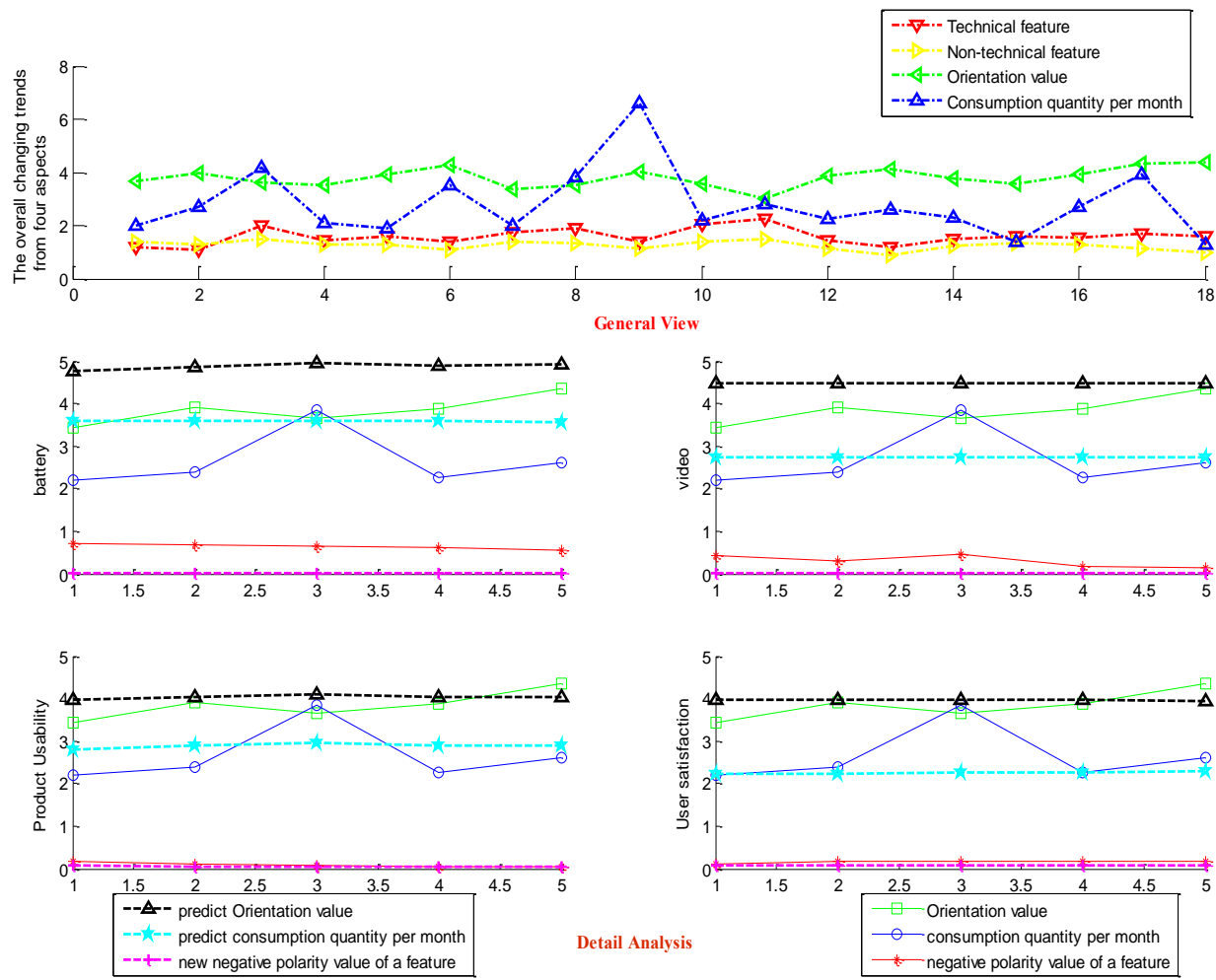

**Figure 8.** Product SX280 HS: Orientation and consumption quantity changing trend based on feature-varying changing trend

**Table 13.** Extracted number of opinion words comparison regarding different size of opinion lexicon

| Product ID | Dictionary Range | avg. num. opinion extracted per review | | avg. (new extracted per review) **Our method** | | | avg. (new extracted per review) **Qiu2011** | |
|---|---|---|---|---|---|---|---|---|
| | | POS words | Neg words | POS words | Neg words | Intensity words | POS words | Neg words |
| S110 | 10% | 0.0830 | 0.1287 | 0.2239 | 0.3826 | 3.5726 | 0.1776 | 0.3281 |
| | 20% | 0.4706 | 0.3380 | 1.3761 | 1.5894 | 3.5726 | 1.008 | 1.4591 |
| | 50% | 1.4267 | 1.0771 | 1.5667 | 1.5747 | 3.5726 | 1.3767 | 1.5524 |
| | 80% | 2.1667 | 1.8138 | 2.7329 | 2.0160 | 3.5726 | 2.5917 | 1.9614 |
| | 100% | 2.7613 | 2.0962 | **3.3328** | **3.1760** | **3.5726** | 3.006 | 3.1621 |
| SX510 HS | 10% | 0.1017 | 0.1405 | 0.2069 | 0.2577 | 3.1578 | 0.1237 | 0.2473 |
| | 20% | 0.6503 | 0.5303 | 1.2426 | 1.1397 | 3.1578 | 1.2411 | 1.1324 |

| | | | | | | | |
|---|---|---|---|---|---|---|---|
| | 50% | 2.1143 | 1.2801 | 1.8836 | 1.1739 | 3.1578 | 1.6792 | 1.1238 |
| | 80% | 3.1728 | 1.7985 | 2.1124 | 2.2439 | 3.1578 | 2.1153 | 2.2040 |
| | 100% | 4.6684 | 2.4487 | **2.7500** | **2.7412** | **3.1578** | 2.7327 | 2.5341 |
| SX280 HS | 10% | 0.0782 | 0.2220 | 0.3092 | 0.3688 | 2.9785 | 0.2560 | 0.2997 |
| | 20% | 0.5640 | 0.6035 | 1.3196 | 1.1546 | 2.9785 | 1.1818 | 1.0753 |
| | 50% | 2.1636 | 1.4267 | 1.8152 | 1.2680 | 2.9785 | 1.7113 | 1.2245 |
| | 80% | 4.1919 | 3.0513 | 2.9011 | 2.8280 | 2.9785 | 2.7935 | 2.6211 |
| | 100% | 4.2130 | 3.5205 | **3.2336** | **3.1893** | **2.9785** | 3.1329 | 3.1776 |

We do a statistical work for 50 review documents in each dataset. Average number of opinion words in one review with and without duplicated words are counted and shown in Table 14. The average number of extracted opinion words from Qiu2011 is less than the number obtained from our method and therefore we select the obtained results from our method. The sum total of extracted opinion words per review from S110, SX510 HS, and SX280 HS is 12.3020, 12.4854, and 15.9508 respectively when the dictionary range is 80%, which is less than the corresponding average number opinion words per each review. This result reveals that some existing opinion words per each review have not been discovered yet. The sum total of extracted opinion words from three datasets is 14.9389, 15.7661, and 17.1349 respectively, which is nearly equal with the average number of opinion words that should be in one review. Meanwhile, the experimental performance of precision, F-score, and especially recall, when the dictionary range is 80%, is lower than the performance of the full range dictionary. Hence, we recommend using the complete opinion words dictionary to do experiments.

**Table 14.** Average number of opinion words per review (or per sentence) in three datasets

| Product ID | No. Reviews | Sentences | Avg. no. of opinion words per review | | Avg.no. of pinion words in one sentence | |
|---|---|---|---|---|---|---|
| | | | Include Duplicated | Not include | Include Duplicated | Not include |
| S110 | 232 | 2054 | 14.6704 | 12.2479 | 1.7611 | 1.3628 |
| SX510 HS | 381 | 2456 | 15.4716 | 12.1250 | 2.7563 | 1.9312 |
| SX280 HS | 517 | 4992 | 17.0519 | 12.7863 | 1.8797 | 1.2583 |

Theoretically, the algorithm does unlimited iterations and propagation between opinion words and features, which can completely extract the information no matter how small the opinion dictionary is. Conversely, the algorithm loses the ability to provide sufficient opinion words and features when the dictionary size is small. We have done a deeper analysis and realized that the following reasons that led to such phenomenon: Firstly, the reviewers usually discuss the same subject in multiple ways. For instance, the subject can be changed into the attributes of the subject, synonyms, or the other aspects of the products. Secondly, the reviewers also adopt different opinion words and different expressive ways to explain the same opinion meaning. The algorithm itself can only iterate and propagate among the same words and therefore a good sized opinion words dictionary is necessary.

### 5.3.2 The superiorities and deficiencies of the proposed algorithm

Performance comparison in section 5.2.1 shows that our method is more efficient than the baseline methods. Our method can also produce more adequate information than Qiu2011 and Jakob2010. We take one sentence from a review document as an example: *"just after few months of use, the camera started to overheat and the battery life was extremely short. And after that time it can't even work."* The extracted information from our method and baseline methods is shown in Table 15.

Qiu2011 and our method have strong dependencies with the opinion dictionary, and Jakob2011 has a strong dependency on the initial labeled training dataset. Qiu2011 obtained the fewest results, because the initial opinion dictionary does not contain any opinion word of the phrase 'battery life was extremely short' and as a result the propagation of the approach won't work with this phrase. Jakob2010 cannot obtain 'short' because of the fact that it is an equivocal word, which has to be determined under special context and it is difficult to estimate during the CRF inference process. Our method can extract more opinion words due to the fact that an additional intensity dictionary is given, and the dependency relations among phrases are also considered. A highlighted advantage is the opinion-feature relations, because it better reflects the real intention of a review document.

Several issues are still unsolved in our work. Our algorithm cannot deal with the weak dependency relations in long sentences and does not have the ability to extract all of the features and opinion words. The proposed method has the ability to report that the reviews have adopted ironic and subjunctive expressions based on the gap of the newly obtained orientation value and the original assigned value. But no strategy is given to calculate the accurate polarity value of ironic and subjunctive expressions in a review text. For the wrong writing word problem, we di-rectly recommend the nearest approximate word to substitute the wrong words based on WordNet dictionary, which is not a deep study. These unsolved issues will be further studied in the future work.

**Table 15.** Extracted information comparison between baseline method and our method

|  | Extracted Opinion words | | Extracted Features | Linked Feature-Opinion |
|---|---|---|---|---|
|  | From Dic | New | -- |  |
| Qiu2011 | *overheat work* | *n't; even* | *camera* | -- |
| Jakob2010 | -- | *overheat; extremely; n't; even; work;* | *camera; battery life* | -- |
| Our Method | *overheat work* | *extremely; short; n't; even* | *camera; battery life* | (camera-overheat)<br>(battery life-extremely short)<br>(it- n 't even work) |

## 6. Conclusions and Future Work

In this paper, we proposed an opinion mining extraction algorithm that can jointly identify features, opinion expressions, and feature-opinion by using fuzzy logic to determine opinion boundaries and adopting syntactic parsing to learn and infer propagation rules between opinions and features. Our algorithm allows opinion extraction to be executed at the phrase level and can automatically detect the features that contain more than one word by building kernels through closest words. This work presents opinion intensifier sets that can aid to extract opinion degree words. In addition, we also have discovered more dependency relations between features and opinions than the previous works. Experimental evaluations show that our algorithm outperforms the baseline approaches on different extraction tasks. In detail, the main advantage of the proposed algorithm is shown as follows:

1) One big difference between our work and previous opinion extraction works (Johansson and Moschitti, 2011; Qiu et al., 2011; Choi and Cardie, 2010; Wu et al., 2009; Hu and Liu, 2006) is that the extracted opinion expression is an opinion unit (phrase level) instead of a single opinion word (token level). Most of the opinion extraction works (Qiu et al., 2011; Brody and Elhadad, 2010; Rill et al., 2012) only adopt the adjective words as opinion expressions, while the opinion unit in our work is considered as a pattern that is composed by adjectives, adverbs, and verbs. We extend the phrase patterns and infer that the pattern should have a maximum of three consecutive words by comparing with the exiting works (Kar and Mandal, 2011; Turney, 2002). An opinion degree intensifier for searching words that modify the intensity of opinion expressions is also proposed.

2) Our method also differs from the existing feature extraction works (Popescu and Etzioni, 2007; Jakob and Gurevych, 2010; Choi and Cardie, 2010). It iteratively builds the feature relations based on the basic semantic meaning (dependency relations) of reviews with the aim of overcoming the problem of the current topic models that cannot always correlate with human perspicacity. The proposed opinion mining algorithm builds a kernel representing the nouns that are closely related at word level, and then a new kernel is continually built to represent the kernels in phrase level that have the close distance.

3) A single opinion word expresses the opinion strength to a certain degree. The opinion expression is the combination of the weights of individual opinion words. Different types of opinion patterns are proposed to organize different combinations. Towards evaluating the vagueness of opinion expressions, we propose fuzzy logic to estimate the polarity intensity of the opinion, quantify the expressed sentiment into a fuzzy number, evaluate the fuzzy weights of product features, and estimate the true opinion strength of reviewers on the corresponding product features.

Two interesting observations are worth being reiterated. The observation through feature-by-feature comparison is that the orientation value of SX280 HS is the lowest one in regards to most reviews frequently criticizing the bad battery dimension, and yet the consumption quantity of this product still maintained a satisfactory level approval rating. This phenomenon reveals that the frequent feature of reviews is not the only factor that influences individual purchase decisions. One possible explanation is that some features are more important than the others and play an important role in personalized purchase decisions. Recognition of important features will be further studied. Another interesting observation is shown in prediction analyses, which exposes the fact that the worst dimension is not always the perfect selection to improve product orientation and consumption quantities. Identification of proper features to improve for both product orientation and consumption quantities will be analyzed deeper in the future work as well.